# Autonomous Ramp Merge Maneuver Based on Reinforcement Learning with Continuous Action Space

Pin Wang*, Ching-Yao Chan

California PATH, University of California, Berkeley, bldg. 454, Richmond Field Station, Richmond, 94804, US

pin_wang@berkeley.edu, cychan@berkeley.edu

*Abstract* – Ramp merging is a critical maneuver for road safety and traffic efficiency. Most of the current automated driving systems developed by multiple automobile manufacturers and suppliers are typically limited to restricted access freeways only. Extending the automated mode to ramp merging zones presents substantial challenges. One is that the automated vehicle needs to incorporate a future objective (e.g. a successful and smooth merge) and optimize a long-term reward that is impacted by subsequent actions when executing the current action. Furthermore, the merging process involves interaction between the merging vehicle and its surrounding vehicles whose behavior may be cooperative or adversarial, leading to distinct merging countermeasures that are crucial to successfully complete the merge. In place of the conventional rule-based approaches, we propose to apply reinforcement learning algorithm on the automated vehicle agent to find an optimal driving policy by maximizing the long-term reward in an interactive driving environment. Most importantly, in contrast to most reinforcement learning applications in which the action space is resolved as discrete, our approach treats the action space as well as the state space as continuous without incurring additional computational costs. Our unique contribution is the design of the Q-function approximation whose format is structured as a quadratic function, by which simple but effective neural networks are used to estimate its coefficients. The results obtained through the implementation of our training platform demonstrate that the vehicle agent is able to learn a safe, smooth and timely merging policy, indicating the effectiveness and practicality of our approach.

## I. INTRODUCTION

Automated vehicles have the potential to reduce traffic accidents and improve traffic efficiency. A number of automakers, high-tech companies, and research agencies are dedicating their efforts to implement and demonstrate partially or highly automated features in modern vehicles, such as the AI-enabled computational platforms for autonomous driving from NVIDIA [1], the Autopilot from Tesla [2], and 'Drive Me' project by Volvo [3]. Fully autonomous vehicles, e.g. Google self-driving car (WAYMO) [4], are also being tested and may be deployed in the near future.

Different levels of automated functions designed for freeways or expressways are well developed and some of them are being or will be introduced in the market soon, such as Level 2 functions (e.g. adaptive cruise control plus lane keeping, etc.) by various automakers. One example is the Super Cruise by General Motors [5]. However, the implementation of autonomous on-ramp merging still presents considerable challenges. One big challenge is that intelligent vehicle agent should take the long-term impacts into consideration when it decides on its current control action (the "long term" in the study is defined to be the completion of a merge process while at any point along the merging maneuver there is a "current" action). In other words, the actions such as accelerating, decelerating, or steering that the ego vehicle takes at the current moment may affect the success or failure of the merge mission. Another challenge is that the merging maneuver is not only based on the merging vehicle's own dynamic state, but dependent on its surrounding vehicles whose actions may be cooperative (e.g. decelerating or changing lane to yield to the merging vehicle) or adversarial (e.g. speeding up to deter the merging vehicle).

The merging process can be handled at relative ease in most cases by experienced human drivers but the algorithms for automated execution of the merge maneuver in a consistently smooth, safe, and reliable manner can become complex. Most previous studies solve the merging problem by assuming some specific rules. For example, Marinescu et al. [6] proposed a slot-based merging algorithm by defining a slot's occupancy status (e.g. free or occupied) based on the information of the mainline vehicles' speed, position, and behavior of acceleration or deceleration. Chen et al. [7] applied a gap acceptance theory and defined some driving rules to model the decision-making process of the on-ramp merge behavior on urban expressways. These rule-based models are conceptually comprehensible but are pragmatically vulnerable due to their inability to adapt to unforeseen situations in the real world.

Reinforcement learning, a machine learning algorithm which trains itself continually through trials and errors [8], has the potential to allow the vehicle agent to learn how to drive under different or previously unencountered situations by training it to build up its pattern recognition capabilities. Reinforcement learning is different from standard supervised learning techniques, which need ground truth as input and output pairs. A reinforcement learning agent learns from past experience and tries to capture the best possible knowledge to find an optimal action given its current state, with the goal of maximizing a long-term reward which is a cumulative effect of the current action on future states.

In our study, we apply reinforcement learning algorithm on the autonomous driving agent to find an optimal merging policy. In a typical reinforcement learning problem, the state space and action space are often treated as discrete, which simplifies the learning process in a finite tabular setting. However, in reality, the vehicle's state and actions (i.e. vehicle dynamics) are continuous. Discretizing them will result in an extremely large unordered set of state/action pairs and render the solution suboptimal. Therefore, finding ways to treat both the state space and action space as continuous is of primary importance, which forms one cornerstone of our research thesis.

The rest of our paper is organized as follows. A literature review of related works is given in the next section, followed by the description of our proposed reinforcement learning algorithm. Then, the training procedure implemented on a simulation platform and the results are presented. Finally, concluding remarks and discussions are given in the closing section.

## II. LITERATURE REVIEW

The application of reinforcement learning has seen significant progress in the field of artificial intelligence in the past decade. Narasimhan et al. [9] employed reinforcement learning for language understanding of text-based games. Li et al. [10] proposed a hybrid reinforcement learning approach to deal with customer relationship management problems in a company, in order to find optimal actions (e.g. sending a catalog, a coupon or a greeting card) on its customers. Google DeepMind [11] has been applied deep reinforcement learning techniques to develop an artificial agent and let it play classic Atari games. The trained agent shows better performance than a professional human by directly learning game policies from high dimensional image inputs.

In recent years, reinforcement learning has been applied in traffic and vehicle control problems. Some studies applied reinforcement learning in ramp metering control to improve traffic efficiency. Fares et al. [12] designed a density control agent based on reinforcement learning to control the vehicles entering the highway from on-ramps. In the study, they define the state space as a three-dimensional space and the action space as a two-action space (i.e. red and green). Yang et al. [13] used basic Q-learning to increase the capacity at the highway-ramp weaving section. The state space was composed by upstream and downstream volumes, and the action space was represented by discrete ramp-merging rates. Some other studies use reinforcement learning for automated vehicle control. Ngai et al. [14] proposed a reinforcement learning multiple-goal framework to solve the overtaking problem of automated vehicles. They used a quantization method to convert continuous sensor state and action space into discrete spaces. The vehicle can accomplish the overtaking task though it cannot always turn to the desired direction accurately due to the discrete steering angles. Yu et al. [15] investigated the use of reinforcement learning to control a simulated car through a browser-based car simulator. They decreased the action space from 9 actions to 3 actions (e.g. faster, faster-plus -left, faster-plus- right) and tested two reward functions. The simulated car can learn turning operations in relatively large sections without going off-road, however it faces challenges in obstacle avoidance. In these studies, the authors use discrete actions to represent the real-world action space which are fundamentally continuous. It has been learned that discretizing action space can simplify the problems and may lead to fast convergence, but it can also result in suboptimal and unrealistic vehicle performance.

Some attempts are made to use continuous action space. Sallab et al. [16] formulated two main reinforcement learning categories, a discrete action category and a continuous action category, for a lane-keeping assistant study. They tested and compared the performance of the two algorithms with an open source car simulator (TORCS), and results showed that discrete action space made steering abrupt while continuous action space gave better performance with smooth control. Shalev-Shwartz et al. [17] applied reinforcement learning to optimize long-term driving strategies (e.g. double merging scenario) where they decomposed the problem into a learnable part and a non-learnable part. The learnable part maps the state into a future trajectory which enables the comfort of driving, while the other part is designed as hard constrains which guarantees the safety of driving. The proposed framework is plausible but the authors has not conducted reproducible experiments.

We believe it is challenging but crucial to consider the control action space as continuous. In our work, we design a unique format of Q-function approximator to obtain the optimal merging policy without increasing computational cost. We give the description of our approach in the next section.

## III. METHODOLOGY

In this section, we provide an in-depth explanation of the methodologies, including the concept of reinforcement learning, the state space, the action space, the reward function, and the neural network based Q-function approximator.

### A. Reinforcement Learning

In a reinforcement learning problem, an agent interacts with the environment which is typically formulized as a Markov Decision Process (MDP). The agent takes the environment observations as state $s \in S$, and chooses an action $a \in A$ based on $s$. After the action execution, it observes the reward $r \sim R(s,a)$ and next state $s' \in S$. An expected discounted cumulative return $G$ is calculated as in (1) based on rewards starting from state s and thereafter following the policy $\pi$. The goal of the reinforcement learning agent is to find an optimal policy $\pi^*$ which maps states into actions.

$$G = E[\sum_k \gamma^{k-1} r_k] \quad (1)$$

where $\gamma$ is a discount factor $\gamma \in (0,1)$.

To solve a reinforcement learning problem, model-based and model-free approaches are two main categories. For the ramp merging problem, it is difficult to prescribe an accurate model of the environment with a state transition matrix. Therefore, we resort to Q-learning, a model-free approach, for finding an optimal driving policy. A Q-function is used to evaluate the long-term return $G(s,a)$ based on the current and next step information $(s,a,r,s')$, instead of waiting until the end of the episode to gather a discounted cumulated reward. $Q(s,a)$ is called the action-state value, among which the highest one $Q^*(s,a^*)$ indicates the action $a^*$ is an optimal action in state $s$. By iteratively updating the estimated Q-values with the observed reward $r$ and next state $s'$ as follows, an optimal policy can be learned.

$$Q(s,a) \coloneqq Q(s,a) + \alpha(r + \gamma * \underset{a'}{argmax}\, Q(s',a') - Q(s,a)) \quad (2)$$

where $\alpha$ is learning rate. An optimal policy ($\pi^*$) is better than or equal to all other policies ($\pi^* \geq \pi, \forall \pi$) in which all the states reach the optimal action values ($Q(s,a) = Q^*(s,a^*)$).

Note that the above update approach only applies to discrete states and actions, which makes it impractical to be applied in our case where both the state space (driving environment) and the action space (vehicle control) are continuous. An alternative is to use neural networks as Q-function approximator. The Q-value for a given state s and a chosen action a is estimated by the Q-network with weights $\theta$, expressed as $Q(s,a,\theta)$. The Q-network can be updated by stochastic gradient descents.

However, if we directly put the states and actions into the neural network without explicitly or implicitly 'tell' it some prior knowledge, it may have a hard time learning the driving policy. Due to this reason, we design the format of the Q-function approximator as a quadratic function to ensure that there is always a global optimal action for a given state at the very moment. The coefficients of the quadratic function are learned by concise neural networks. To setup the learning graph, we first define the state space, action space and reward function, and then formulate the Q-function approximator. These are described in the following sections.

## B. State Space

In a typical on-ramp merging scenario, the ego vehicle (i.e. the merging vehicle) needs to know not only its own dynamic state but also the state of its surrounding vehicles (SVs) to make a rational decision on when and how to merge onto the highway. In other words, the ego vehicle's state is related to SVs' state which makes the driving environment a Non-MDP.

It is a fact that the real-world environment is rarely a MDP, but many situations can be approximated as a MDP in one way or another. In our case, the ego vehicle's own state is independent of its historical kinematic information given its current state (which is a MDP), while the SVs' states are not in the view of the ego vehicle mainly due to the unpredictable nature of their next state (which makes it a Non-MDP). The historical vehicle dynamic information of highway vehicles may give a hint about how they will probably behave in a short future and this can be learned by a LSTM (Long Short Term Memory) based model as we previously proposed in our early work [18], but the most critical information valuable for the ego vehicle to select an optimal action is their current states. Besides, due to the advanced sensing technologies in positioning, communicating, and computing, we can capture the vehicles' state instantaneously (tens to hundreds of milliseconds) and simultaneously transmit it to the agent control modular for the process of perception, recognition and action decision. In this sense, we currently simplify the real-world driving environment into a MDP.

The merging procedure can be partitioned into three phases. First, find an appropriate gap. To do this, the ego vehicle needs to estimate the arrival time to the merging section of its own and of the other vehicles on the highway. Second, execute merging maneuver. The ego vehicle needs to adjust its action to merge safely and smoothly into the selected gap, and this is what the vehicle agent needs to be trained. After completing the merging, the ego vehicle should be able to perform proper car-following actions as vehicles on the highway usually do. In the overall process, the dynamics of the gap-front vehicle (meaning the vehicle directly ahead of the ego-vehicle) and the gap-back vehicle (meaning the vehicle directly behind) are critical for the ego vehicle to learn the optimal merging policy. Thereby, the state space is defined to include the dynamics of the ego vehicle, the gap-front vehicle and the gap-back vehicle. Additionally, we add another element, the highway speed limit, to constrain the vehicle's speed in a reasonable range. The continuous state space is therefore defined as

$$s = (v_{ev}, p_{ev}, v_{gfv}, p_{gfv}, v_{gbv}, p_{gbv})$$

where $v_{ev}$ and $p_{ev}$ are the speed and position of the ego vehicle; $v_{gfv}$ and $p_{gfv}$ are the speed and position of the gap front vehicle; $v_{gbv}$ and $p_{gbv}$ are the speed and position of the gap back vehicle.

## C. Action Space

Typically, vehicle control refers to longitudinal control (e.g. acceleration or deceleration) and lateral control (e.g. steering). In the on-ramp merge scenario, we suppose the merging vehicle travels along the centerline of the lane from ramp to highway and such geometry information is available from the embedded digital map for the ego-vehicle to follow. In other words, for the purpose of demonstrating the reinforcement learning concept in this paper, we do not include the lateral control of steering and only model the longitudinal acceleration as the control action.

Based on vehicle dynamics it is common sense that in reality the acceleration of a vehicle cannot be an arbitrarily value. Therefore, we limit the acceleration in a range of [-$4.5 m^2/s$, $2.5\ m^2/s$] based on literature on vehicle dynamics [19], and allow the acceleration to be any real value within the range, which is different from some other studies in which the acceleration space was divided into some subsets or a sequence of discrete numbers.

It is worth mentioning that the output action from the learning algorithm generally takes effect on the agent for a relatively small time interval when the data update frequency is high (e.g. 10Hz), leading to miniscule or unobservable effects of that action. To overcome this phenomenon, in evaluating the vehicle dynamics we keep the action the same for a few steps (e.g. the next $k$ steps) to let it manifest its impact, and then update it based on newly observed information. In other words, the action calculation is updated every $k$ steps, while the state is updated at every time step.

## D. Reward Function

After the reinforcement learning agent takes an action in a given state, its impact on the environment is fed back as an immediate reward, i.e., the immediate reward measures the effect of an action in a given state.

In our on-ramp merging problem, the effect is reflected by the smoothness, safeness, and promptness of the merging maneuver. Smoothness represents the comfort of the merging maneuver and is measured by the absolute value of the acceleration. The higher the absolute value of the acceleration is, the larger penalty will be imposed on the agent. The safeness is estimated by the distance to the surrounding vehicles. The closer the ego vehicle is positioned to its surrounding vehicles, the larger the penalty it gets. The promptness is assessed by the time that the ego vehicle will take to complete the merging process. This effect cannot be immediately measured by only a single time interval since merging is a time sequential process. We resort to the current vehicle speed to account for the contribution of promptness in the immediate reward. Consequently, the composition of the immediate reward is expressed in equations (3) - (6).

$$R(s,a) = R_1(acceleration) + R_2(distance) + R_3(speed) \quad (3)$$

$$R_1(acceleration) = f_1 * abs(acceleration) \quad (4)$$

$$R_2(distance) = f_2 * g_2(distance) \quad (5)$$

$$R_3(speed) = f_3 * speed \quad (6)$$

where $f_1$, $f_2$, and $f_3$ are factors accounted for each part of reward.

It needs to be stressed that the importance of the safeness is relatively higher than the smoothness and timeliness in our daily driving. Hence we put more emphasis on the distance related reward. This reward is split two parts, the reward from the distance to the gap front vehicle and the reward from the distance to the gap back vehicle, respectively. Equation (5) is further specified as

$$R_2(distance) = f_{21} * g_{21}(dis_{gfv}) + f_{22} * g_{22}(dis_{gbv}) \quad (7)$$

where $g_{21}$ and $g_{22}$ are functions of the distance to gap-front vehicle ($dis_{gfv}$) and the distance to gap-back vehicle ($dis_{gbv}$), and $f_{21}$ and $f_{22}$ are the corresponding factors respectively. Note that when the ego vehicle is far from or has passed the merging zone, $dis_{gfv}$ and $dis_{gbv}$ are not necessarily important to the ego vehicle, therefore, $f_{21}$ and $f_{22}$ are set to zeros when the ego vehicle is relatively far from the merging zone.

The factor $f_1$ for the acceleration in the immediate reward function is relatively straightforward and can be assigned as a constant. The speed factor $f_3$ depends on how fast a merging behavior is considered appropriate and acceptable, and can be designated as a polygonal function to punish speed values that are too low or too high.

### E. Q-function Approximator

The quadratic format of Q-function approximator is specified as follows

$$Q(s,a) = A(s) * (B(s) - a)^2 + C(s) \quad (8)$$

where $A$, $B$, and $C$ are trainable parameters and designed with the neural network structure with environment state as inputs. An illustration is shown in Fig. 1.

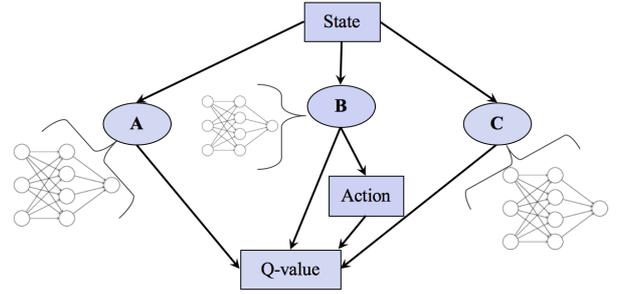

Fig. 1. Graph of the Q-function approximator.

There are two graphs concealed in this form of the Q-function approximator. One is the graph for obtaining an optimal action in a given state, the other is the graph for calculating the Q-value for a given state and action. In the optimal action graph, the optimal action is obtained as $a^* = B(s)$, where $B(s)$ is learned based on the current state $s$. In the Q-value graph, the Q-value is calculated based on the coefficients $A(s)$, $B(s)$, $C(s)$, and action $a$, where the coefficients are constructed by neural networks with the state $s$ as fundamental input. $A$ is assigned a negative value with an activation function used in the neural network. $B$ has the same structure as that in the optimal action graph.

In the learning process, Q-network is updated with the following loss function.

$$Loss = \sum_{i=1}^{n}(r + \gamma * \max_{a'} Q(s',a',\theta) - Q(s,a,\theta))^2_i \quad (9)$$

where $r + \gamma * \max_{a'} Q(s',a',\theta)$ is called the target Q-value and $Q(s,a,\theta)$ is called the predicted Q-value in our manuscript. $\theta$ is a set of Q-network parameters.

When the agent is trained based on equation (9), stability issues and correlations in the observed sequence are factors affecting the learning performance. Experience replay and a second Q-network are good techniques to alleviate the problem [20]. For experience replay in our research, a mini-batch of training samples $(s,a,r,s')_i$ are selected from a replay memory and fed into the learning graph. For each sample tuple $(s, a, r, s')$, $s$ is taken as input to the neural networks of $A$, $B$, and $C$ to obtain their values, and at the same time $a$ is also input to the Q-function approximator to obtain $Q_P$. The calculation of $Q_T$ is a combined process of the optimal action calculation and Q-value calculation. It first calculates the optimal action $a'$ based on the state $s'$ with the use of the optimal action graph, then it calculates the Q-value $Q(s', a')$ by using the Q-value graph with the inputs of next state $s'$ and the optimal action $a'$.

To break the correlations, a second Q-network, called the target Q-network, which has the same structure but different parameter values ($\theta^-$) with the original Q-network ($\theta$), called the prediction Q-network, is used to calculate the target Q-values. The loss, expressed as the summed errors between the predicted Q-values $Q_P(s, a, \theta)$ and the target Q-values $Q_T(s, a, \theta^-)$, is rewritten as follows

$$Loss = \sum_{i=1}^{n}(r + \gamma * \max_{a'} Q(s', a', \theta^-) - Q(s, a, \theta))_i^2 \quad (10)$$

where $\theta^-$ is the parameters in the target Q-network, and $\theta$ is the parameters in the prediction Q-network. In the learning process, $\theta$ is updated at every time step while $\theta^-$ is updated periodically. A step-by-step learning procedure is shown in Fig. 2.

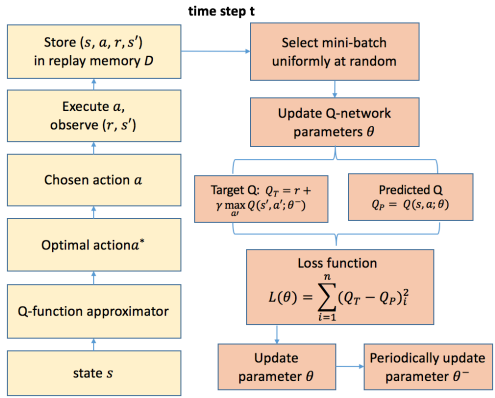

Fig. 2. Reinforcement learning procedure.

## IV. SIMULATION AND RESULTS

### A. Simulation Settings

We train our reinforcement learning agent in simulated ramp merging scenarios where the ramp is a 3.5m wide lane and the main highway is a two-way four-lane highway with a lane width of 3.75m. The highway speed limit is 65 mi/h. The highway traffic is composed of randomly emerging vehicles with random initial speed at the entrance of the highway section, and the highway vehicle can perform car following behaviors when it is close to its leading vehicle. More importantly, the highway vehicles can yield to or surpass the ego vehicle when the ego vehicle is about to merge onto the highway, representing the real-world cooperative or adversarial situations. On the ramp, there is always one ramp vehicle (i.e. the ego vehicle) travelling towards the highway. After one ramp vehicle completes its merging task, another ramp vehicle departs at the beginning of the ramp and is the new ego vehicle.

The ego vehicle is supposed to be equipped with a suite of sensors including lidar, radar, camera, a digital map, DGPS (Differential Global Positioning System) and IMU (Inertial Measurement Unit), and can gather the vehicle dynamic information of its own and its surrounding vehicles within a vicinity of 150m that is also assumed to be accurate enough to meet our requirements. These assumptions are far from the realistic situations where the observation range may be partially occluded and the measurements are shortened, imprecise or inaccurate. Within the scope of this paper, we leave the assumptions alone, but in future work the sensing capabilities of the ego-vehicle can be adjusted to represent various scenarios and measurement conditions.

### B. Training

The training procedure is illustrated as follows in Table 1.

TABLE 1 Training Procedure

| |
|---|
| 1  Initialize experience replay memory $D$ |
| 2  Initialize neural network parameters $\theta$ |
| 3  Initialize the environment state $s$ |
| 4  Set values for $N, k, M, p, dt, \gamma, \alpha$ |
| 5  for step $i = 1: N$ do |
| 6      $t = t + dt$ |
| 7      if $i$ mod $k == 0$ do |
| 8          get the current state $s$ |
| 9          calculate the best action $a^*$ based on $s$ with optimal action graph |
| 10         get the chosen action $a$ with random noise |
| 11         execute $a$ |
| 12         obtain immediate reward $r$ and next state $s'$ |
| 13         store the transition tuple $(s, a, r, s')$ into $D$ |
| 14     sample mini-batch $e_i = (s_i, a_i, r_i, s_{i+1})$ from $D$ ($\forall i \in (1, M)$) |
| 15     for $j = 1$ to $M$, do |
| 16         calculate predicted Q-value: $Q_j^P = Q(s, a, \theta)_j$ |
| 17         calculate target Q-value: $Q_j^T = r_j + \gamma * max_{a'} Q(s', a'; \theta^-)_j$ |
| 18         calculate loss: $Loss = \sum_{j=1}^{M}(Q_j^T - Q_j^P)_j^2$ |
| 19         update network parameters by backpropagation with learning rate $\alpha$ |
| 20     if $i$ mod $p == 0$ do |
| 21         update $\theta^-$ with $\theta$: $\theta^- = \theta$ |

In our study, we design the neural networks in $A$, $B$, and $C$ with a two-layer neural network. The total training steps $N$ are set to 1,600,000, during which there are around 8,000 ramp vehicles performed ramp merging behavior. The data update interval $dt$ is set to 0.1s. The action update step $k$ is set to 4. The size of replay mini-batch $M$ is set to 32. The target Q parameter update step $p$ is set to 500. The discount factor $\gamma$ in the calculation of $Q_T$ is set to 0.95. The learning rate $\alpha$ in the backpropagation is set to 0.001.

*C. Results*

The loss calculated based on $Q_P$ and $Q_T$ is plotted along with the training steps in Fig. 3. To save computation memory, loss values are store every 5 steps. The graph shows an obvious decaying and converging trend despite a few spikes along the way. It is normal to have some spikes as in daily driving one may encounter some extreme situations where an unusual action such as a hard braking is required.

We also accumulated the immediate rewards for each ramp merging vehicle in a complete merging task.

Fig. 4 shows the total reward (named single total reward) of all the 8000 vehicles in the simulation. To be specific, each point on the curve is a cumulative result of the immediate rewards that the ramp vehicle obtains at each time step during its merging process. Note that the values are always be negative since the immediate rewards are defined as a penalty whose value is always negative by our definition.

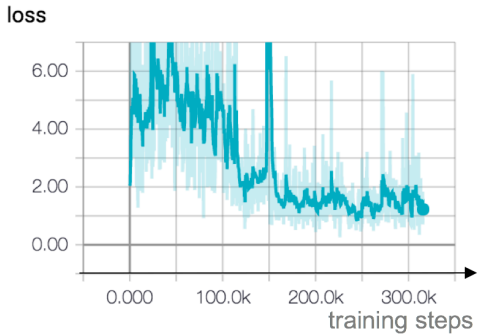

Fig. 3.   Training loss curve.

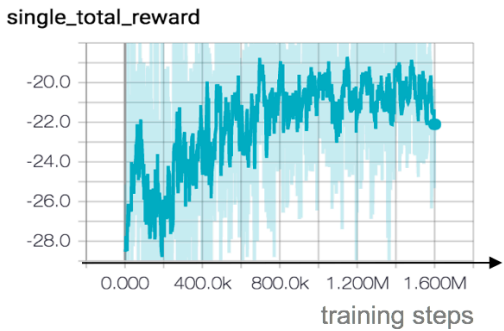

Fig. 4.   Curve of single total rewards of ramp vehicles.

Remember that the total reward is composed of four parts, reward from distance to front vehicle, reward from distance to gap back vehicle, reward from acceleration, and reward from speed. We also plot the four curves along with the training steps in Fig. 5.

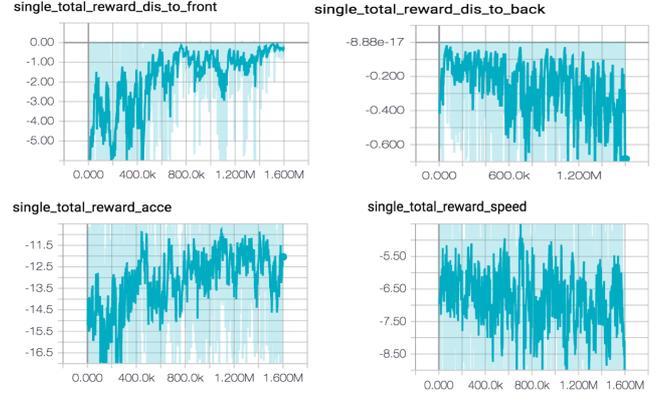

Fig. 5.   Individual rewards from total reward.

From Fig. 5 we can see that the reward curves of distance to gap-front vehicle (single_total_reward_dis_to_front) and vehicle acceleration (single_total_reward_acce) show apparent convergence. In these two graphs, the rewards go up from large negative values to relatively small values and show a potential steady trend, similar with the single total reward curve. In contrast, reward curves of the distance to gap-back vehicle (single_total_reward_dis_to_back) and the acceleration (single_total_reward_speed) show a higher level of fluctuations. The explanation for the curve of single_total_reward_dis_to_back is that we put greater emphasis on the front safety in our design, so the vehicle agent will learn to try to keep relatively large distance to the preceding vehicle while compromising the distance to the gap-back vehicle. Another reason is that the distance to the gap-back vehicle is not entirely controlled by the ramp merge vehicle as it is also affected by the action of the gap-back vehicle. As for the fluctuation of the speed reward curve, it is more intuitional to understand because the speed is adjusted to accommodate to the smoothness and safety purpose and it has the least weight compared to the other three parts in the reward function.

## V.   CONCLUSION AND DISCUSSION

In this work, we adopted a reinforcement learning approach for developing an on-ramp merge driving policy. Our key contribution is that we treat the state space and action space as continuous as in the real-world situation, in order to learn a practical automated control policy. The reward function is designed based on intuitive concerns of human drivers in a merging situation where safeness, smoothness and promptness are the primary attributes reflecting the success of a merge maneuver. It is formulated with vehicle acceleration, speed, and distance to gap vehicles, which are all explicit variables and can directly measure the performance of the merging maneuvers. Another contribution of our work is the unique format of the proposed Q-function approximator that guarantees the existence of an optimal action in a given state without complicating the neural networks' structure. The training results show that the automated vehicle agent is able to learn to merge safely, smoothly and timely onto the highway as the training goes on for a period of time which indicates the validity of our methodology.

There is still room to further improve the performance of the learning agent. One aspect is to fine-tune the reinforcement learning model by trying different hyperparameters, for example the structure of the neural networks, the weights update frequency of the target Q-network, etc. Besides, proper state feature engineering and different reward function compositions are additional promising points worth investigating in the future.